\documentclass[runningheads]{llncs}
\usepackage{amssymb}
\usepackage{graphicx}
\begin{document}
\title{COTR: Convolution in Transformer Network for End to End Polyp Detection}
%
%
\author{Zhiqiang Shen\inst{1} \and
Chaonan Lin\inst{1} \and
Shaohua Zheng\inst{1}}

\authorrunning{Zhiqiang et al.}
%
\institute{Fuzhou University \\
\email{sunphen@fzu.edu.cn}}
\maketitle              
\begin{abstract}
Purpose: Colorectal cancer (CRC) is the second most common cause of cancer mortality worldwide. Colonoscopy is a widely used technique for colon screening and polyp lesions diagnosis. Nevertheless, manual screening using colonoscopy suffers from a substantial miss rate of polyps and is an overwhelming burden for endoscopists. Computer-aided diagnosis (CAD) for polyp detection has the potential to reduce human error and human burden. However, current polyp detection methods based on object detection framework need many handcrafted pre-processing and post-processing operations or user guidance that require domain-specific knowledge. 
\\
Methods: In this paper, we propose a convolution in transformer (COTR) network for end-to-end polyp detection. Motivated by the detection transformer (DETR), COTR is constituted by a CNN for feature extraction, transformer encoder layers interleaved with convolutional layers for feature encoding and recalibration, transformer decoder layers for object querying, and a feed-forward network for detection prediction. Considering the slow convergence of DETR, COTR embeds convolution layers into transformer encoder for feature reconstruction and convergence acceleration. 
\\
Results: Experimental results on two public polyp datasets show that COTR achieved 91.49\% precision, 82.69\% sensitivity, and 86.87\% F1-score on the ETIS-LARIB, and 91.67\% precision, 93.54\% sensitivity, and 92.60\% F1-score on the CVC-ColonDB.
\\
Conclusion: This study proposed an end to end detection method based on detection transformer for colorectal polyp detection. Experimental results on ETIS-LARIB and CVC-ColonDB dataset demonstrated that the proposed model achieved comparable performance against state-of-the-art methods.

\keywords{Colorectal cancer  \and Colonoscopy \and Polyp detection \and Transformer.}
\end{abstract}
\section{Introduction}
\label{sec:introduction}
Colorectal cancer (CRC) is the second most common cause of cancer mortality worldwide \cite{GlobalCancerStatistics2020}. More than 1.9 million new CRC cases and 935,000 deaths were estimated to occur in 2020, representing about one in 10 cancer cases and deaths \cite{GlobalCancerStatistics2020}. Most CRCs derive from adenomatous polyps, initially benign growth on the inner lining of the colon and rectum. However, polyps may become malignant over time and spread to nearby organs due to late diagnosis. The 5-year relative survival rate for CRC ranges from 90\% for patients diagnosed with early-stage to 14\% for those diagnosed with distant-stage disease \cite{ColorectalCancerStatistics2020}. Therefore, early detection and removal of colorectal polyps are vital for survival. Colonoscopy is a widely used technique for colon screening and polyp detection during clinical procedures \cite{PolypSyndrome}. However, the detection accuracy highly depends on the experience of endoscopists. Recent clinical studies have shown that around 25\% of polyps are missed on the average per subject \cite{leufkens2012factors}. On the other hand, the screening of enormous visual information of colonoscopy is a burden for endoscopists. Computer-aided diagnosis (CAD) acting as the second observer has the potential to improve the polyp detection rate and reduce the burden of doctors.
Over the last two decades, various CAD systems have been developed for polyp detection. Traditionally, the majority of approaches toward polyp detection are based on the extraction and classification of manually designed features, such as wavelet transform-based features \cite{karkanis2003computer,iakovidis2006intelligent}, edge, texture, and shape-based approaches \cite{bernal2012towards,ameling2009texture,tajbakhsh2015automated}. In \cite{bae2015polyp} an imbalanced learning scheme with a discriminative feature learning was proposed to address balance training between polyp and non-polyp images. In \cite{deeba2020computer} a saliency-based selection method with the histogram of oriented gradient features was proposed for polyp detection. However, these feature patterns often subject to poor performance in polyp and polyp-like normal structures result in the limitation of feature expression. Recently, deep learning-based detection models that adopted convolutional neural networks (CNN) have been successfully applied for automatic polyp detection \cite{bernal2017comparative,shin2018automatic,qadir2019improving,qadir2021toward}. Shin et al. adopted region-based two-stage CNN for polyp candidates screening and subsequently post-learning methods for false-positive reduction \cite{shin2018automatic}. Similarly, Qadir et al. proposed a polyp detection framework consisting of two stages: a region of interest proposal by a CNN-based object detector and a false positive reduction component \cite{qadir2019improving}. In \cite{qadir2021toward}, the author applied an encoder-decoder net-work to predict Gaussian masks. Then, additional post-processing operations were employed to calculate polyp locations and confidences. However, the region-based polyp detection methods need anchors presetting to encode the prior knowledge and non-maximum suppression to refine the detection results. These pre and post-processing steps significantly influence the detection performance.
To develop an end-to-end CAD system, in this paper, we propose a convolution in transformer (COTR) network for polyp detection without the need for additional pre or post-processing operations. COTR is composed of a CNN for feature extraction, transformer encoder layers interleaved with convolutional layers for feature encoding and reconstruction, transformer decoder layers for object prediction, and a feed-forward network for detection prediction. The ethos of COTR is derived from the detection transformer (DETR) that is an end-to-end detector in natural image scenarios \cite{DETR}. However, DETR often suffers from slow convergence. We speculate that the convergence problem is attributed to the image feature structure is disorganized by the transformer encoders that regard image features as a sequence. Since the training data scale of polyp datasets is substantially less than that of natural images and the additional complexity in polyp region detection for its diversified textures, shapes, and colors while having minute differences with the background, the convergence problem will be further aggravated. Therefore, we embed convolutional layers into the transformer encoder for high-level image features reconstruction and convergence acceleration. Experimental results on two public polyp datasets demonstrate the effectiveness of the proposed methods. COTR achieved 91.49\% precision, 82.69\% sensitivity, and 86.87\% F1-score on the ETIS-LARIB, and 91.67\% precision, 93.54\% sensitivity, and 92.60\% F1-score on the CVC-ColonDB. Our contributions can be summarized as follow:
1) To the best of our knowledge, we are the first to apply transformers for end-to-end polyp detection.
2) We insert convolutional layers into transformer encoders to reconstruct the high-level image features and alleviate the convergence problem of transformers in object detection.
3) Experimental results on two public polyp datasets demonstrate the effectiveness of the proposed COTR. Our method achieves comparable performance against state-of-the-art polyp detection approaches.

\begin{figure}
\includegraphics[width=\textwidth]{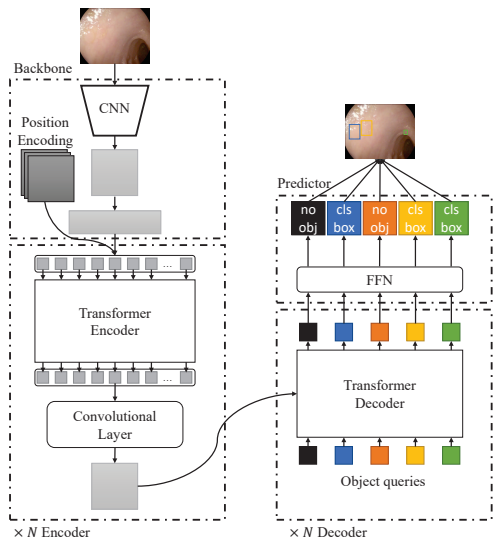}
\caption{Overview of the proposed COTR.} 
\label{fig1}
\end{figure}

\section{Method}
\label{sec:method}
Motivated by the DETR \cite{DETR}, COTR views polyp detection as a direct set predic-tion problem. The proposed end-to-end polyp detector contains two ingredients: the COTR architecture for polyp detection and the bipartite matching loss to force COTR to generate unique predictions. The overview of COTR is illustrated in Fig.\ref{fig1}. COTR architecture will be detail in Section \ref{subsec:convolution_in_transformer_network}. The loss function that transforms objection detection problem into set prediction one will be discussed in Section \ref{subsec:loss_function}.
\subsection{Convolution in transformer network}
\label{subsec:convolution_in_transformer_network}
COTR is consists of a CNN backbone for feature extraction, six convolution-in-transformer encoders for feature encoding and reconstruction, six transformer decoders for object querying, and a simple feed-forward network for detection prediction.

\textbf{Feature Extractor}. We adopt ResNet18 \cite{resnet} as a feature extractor backbone. It takes as input a colonoscopy image $x \in \mathbb{R}^{3\times H_0\times W_0}$ and outputs a high-level feature map $h \in \mathbb{R}^{C\times H\times W}$ where $C = 512$ and $H, W = \frac{H}{32}, \frac{W}{32}$. Moreover, a $1\times 1$ convolution layer with a reduction rate of r is employed to reduces the channel dimension of the high-level activation map h from C to C/r, which results in a new feature map $f \in \mathbb{R}^{\frac{C}{r}\times H\times W}$. 

\textbf{Encoder}. The COTR model contains six convolution-in-transformer encoders. Each COTR encoder consists of a transformer encoder that collapses the spatial structure of f into a sequence $s \in \mathbb{R}^{\frac{C}{r}\times HW}$ and a convolution layer that reconstructs the sequence structure back to the spatial one. Each transformer encoder has a standard architecture and consists of a multi-head self-attention module and a feed-forward network. Positional embedding is introduced to the input of each attention layer \cite{transformer}. Each convolution layer is composed of a $3\times 3$ convolution, a batch normalization, and a rectified linear unit. 

\textbf{Decoder}. The COTR model contains six transformer encoders. Each COTR decoder follows the standard architecture of the transformer except that it decodes object queries in parallel, following the DETR \cite{DETR}. The decoders take as input N object queries with position embeddings and output embeddings of the encoders, and transform them into decoded embeddings which are then conveyed to feed-forward layers for object prediction. 

\textbf{Object Predictor}. The object predictor is a feed-forward network with two sibling fully connected layers, i.e., a box regression layer to predict object (polyp) location (x, y, w, h) and a box-classification layer to predict object (polyp and background) scores. Therefore, N object queries are independently decoded into box coordinates and class labels by the feed-forward network, which results in N final predictions including object (polyp) and no object (background) predictions.
\subsection{Loss function}
\label{subsec:loss_function}
To transform object detection as a direct set prediction problem, we adopt bipartite matching loss to search an optimal bipartite matching between predictions with a fixed size of N and ground truth objects and then optimize object-specific losses. Following the DETR, we employ Hungarian algorithm to search the optimal permutation $\omega \in \Omega_N$.An optimal permutation is the one with the lowest matching cost towards the ground truth, which can be defined as 
\begin{equation}
\hat{\omega}=\mathop{\arg \min }_{\omega \in \Omega_{N}} \sum_{i=1}^{N} L_{match}\left(y_{i}, \hat{y}_{\omega(i)}\right)
\end{equation}
where $y_i$ is an element of the ground truth and $\hat{y}_{\omega(i)}$ is an element of the predictions.Each element of the ground truth objects is denoted as $y_i=(c_i,b_i )$ where $c_i$ is the target class label (polyp and non-polyp), and $b_i$ is the normalized ground truth boxes with center coordinates; each element of the predicted objects refers to  $\hat{y}_{\omega(i)}=(\hat{p}_{\omega(i)}(c_i), \hat{b}_\omega(i))$, where $\hat{p}_{\omega(i)}(c_i)$ is the predicted probability of class $c_i$, and $\hat{b}_\omega(i)$ is the predicted object box.we find a one-to-one matching of class labels and object boxes between the ground truth and the predictions. The loss function for training our network is a linear combination of the classification loss and the localization loss, which is defined as
\begin{equation}
L(\hat{y}, y)=\sum_{i=1}^{N}\left[\alpha_{1} L_{c l s}\left(c_{i}, \hat{p}_{\omega(i)}\left(c_{i}\right)\right)+\alpha_{2} L_{l o c}\left(b_{i}, \hat{b}_{\omega_{(i)}}\right)\right]
\end{equation}
where the classification loss denotes
\begin{equation}
L_{cls}\left(c_{i}, \hat{p}_{\omega(i)}(c_{i})\right)=\sum_{i=1}^{N}-\log \hat{p}_{\omega(i)}(c_{i})
\end{equation}
and the localization loss denotes
\begin{equation}
L_{loc}\left(b_{i}, \hat{b}_{\omega_{(i)}}\right)=\sum_{i=1}^{N}\left[\beta_{1} L_{iou}\left(b_{i}, \hat{b}_{\omega_{(i)}}\right)+\beta_{2} L_{reg}\left(b_{i}, \hat{b}_{\omega_{(i)}}\right)\right]
\end{equation}
To alleviate the loss balance between large and small objects, we employ generalized intersection over union (GIOU) loss \cite{GIOU} and $L1$ loss as $L_{iou}$ and $L_{reg}$, respectively. The $\alpha_1$, $\alpha_2$, $\beta_1$, and $\beta_2$ are hyperparameters that are set as $\alpha_1=1$, $\alpha_2=5$, $\beta_1=2$, and $\beta_2=5$, respectively.

\section{Experiments}
\label{sec:experiments}
\subsection{Experimental datasets}
\label{subset:experimental_datasets}
In this study, we used three publicly available datasets including CVC-ClinicDB for training and ETIS-LARIB and CVC-ColonDB for testing. These datasets include colonoscopy images with the corresponding binary masks manually annotated ground truth covering the polyp. Details of these datasets are presented in the following.

\textbf{CVC-ClinicDB} \cite{bernal2015wm,fernandez2016exploring}: This dataset contains 31 unique polyps extracted from 29 colonoscopy videos and presented 646 times in 612 still images with a pixel resolution of $384\times 288$ in standard definition. 

\textbf{ETIS-LARIB} \cite{silva2014toward}: This dataset consists of 196 still images extracted from 34 colonoscopy videos. The images have a high definition resolution of $1225\times 966$ pixels. Some images contain two or three polyps, making the total number of polyp appearances 208.

\textbf{CVC-ColonDB} \cite{bernal2012towards}: This dataset has 300 still images extracted from 15 video sequences. A random sample of 20 frames per sequence was obtained, with a standard definition resolution of $574\times 500$ pixels. In every image, there exists only one polyp.

\subsection{Evaluation metrics}
\label{subset:evaluation_metrics}
To evaluate the performance of the proposed method and fairly compare it with state-of-the-art approaches, we calculate the following metrics.

\textbf{Precision}.It computes the percentage of the correct detection outputs among all the predicted outputs.
\begin{equation}
{Precision}(Pre)=\frac{TP}{TP+FP}
\end{equation}

\textbf{Sensitivity}. It measures the percentage of the truly detected polyps among all the polyps. 
\begin{equation}
{Sensitivity}(Sen)=\frac{TP}{TP+FN}
\end{equation}

\textbf{F1-score}. It can be used to consider the balance between sensitivity and precision.
\begin{equation}
{F1-score}(F1)=\frac{2\times Pre\times Sen}{Pre+Sen}
\end{equation}

A polyp will be considered to be detected if the centroid of the predicted region falls inside the ground truth region. In case of having multiple true detection out-puts for the same polyp, they will be counted as one true positive (TP). A detect-ed object with a centroid that falls outside the ground truth region will be counted as a false positive (FP). A false negative (FN) is a polyp that is not detected. 

\subsection{Experimental setting}
\label{subsec:experimental_setting}
The experiments have been performed on PyTorch \cite{paszke2019pytorch}. We trained COTR with AdamW \cite{adamW} optimizer setting the initial transformer’s learning rate to $le-4$, the backbone’s to $1e-5$, and weight decay to $le-4$ after 200 epochs on the total number of epochs of 300. We used batch size of 4 due to the GPU memory constraint. To increase the training samples, data augmentation such as flip, crop, and scale have been leveraged. The networks were trained on a single NVIDIA GeForce GTX 1080 with GPU memory of 8G.

\begin{figure}[ht]
\includegraphics[width=\textwidth]{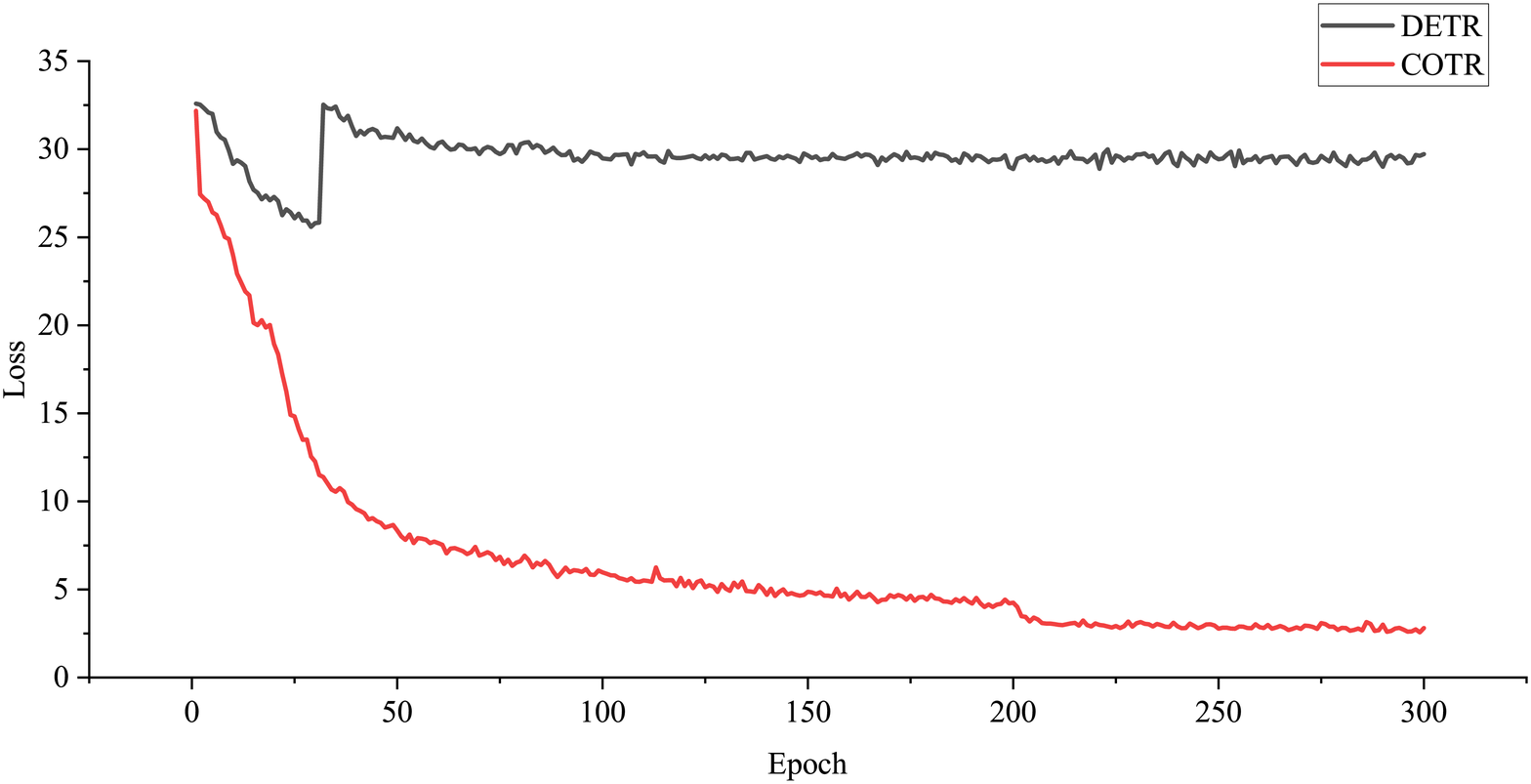}
\caption{Training loss curves of DETR and COTR on CVC-ClinicDB.} 
\label{fig2}
\end{figure}

\begin{figure}[ht]
\includegraphics[width=\textwidth]{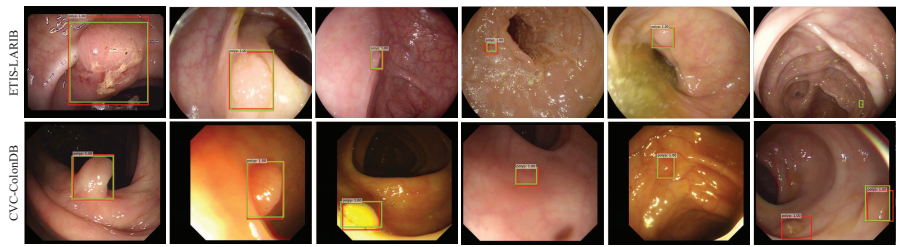}
\caption{Examples of polyps predicted by the proposed COTR. The first row is the results on ETIS-LARIB dataset and the second row is the results on CVC-ColonDB dataset. The green rec-tangles are the ground truth bounding boxes. The red rectangles are the predicted bounding boxes of the COTR.} 
\label{fig3}
\end{figure}

\section{Results}
\label{sec:results}
In this study, we trained our proposed method on CVC-ClinicDB dataset and evaluated it on ETIS-LARIB and CVC-ColonDB datasets, respectively. In the following, we detail the model convergence problem during the training stage and performance comparison with state-of-the-art methods.
\subsection{Convergence comparison on CVC-ClinicDB}
In this experiment, we trained the DETR and COTR on CVC-ClinicDB dataset and compared their convergence situation to investigate the effectiveness of the pro-posed method and validate our speculation that the image structure is disorganized by the transformer encoders. Fig. \ref{fig2} depicts the training losses of both DETR and COTR. The loss plot of DETR remains stubbornly high whereas that of COTR converges to a low level. We would like to intuitively discuss the reason behind the superior performance of our proposed algorithm. The high-level image features extracted by the backbone are disorganized by the transformer en-coders. The convolution layers of COTR embedded in the transformer encoders reconstruct the feature structure, therefore, facilitating the model convergence process.

\subsection{Performance comparison on ETIS-LARIB}
In Table \ref{tab1}, we compare the performance of the proposed COTR against several state-of-the-art models on ETIS-LARIB dataset. As shown, COTR outperformed the other methods in terms of precision and F1 score. Although the sensitivity of COTR lower than that of MDeNetplus, the balance between precision and sensitivity of COTR slightly outperformed that of MDeNetplus. COTR was able to detect polyps with various textures, shapes, and colors, even when they are surrounded by the residual liquid in the large intestine. However, it might be uncertain to detect sessile polyps as true positives. Fig. \ref{fig3} (the first row) shows the predictions of COTR on ETIS-LARIB dataset. 

\subsection{Performance comparison on CVC-ColonDB}
In the following, we further compared the proposed COTR with other methods on CVC-ColonDB. As shown in Table \ref{tab2}, COTR achieved the highest precision (91.67), sensitivity (93.54), and F1-score (92.60) against other approaches. COTR could correctly detect the small polyps. However, it might be disturbed by the hot spot. Examples are shown in Fig. \ref{fig3} (the second row). Considering that the DETR comes with a challenge to identify small objects, we surmise that the convolution layers inserted in the transformer encoders enlarged the receptive field of COTR, thus improved the robustness of the model to detect small polyps.

\begin{table}
\centering
\caption{Performance comparison with state-of-the-art methods on ETIS-LARIB.}
\label{tab1}
\begin{tabular}{lllllll}
\hline
Method               & TP           & FP          & FN          & Pre            & Sen            & F1             \\ \hline
WE-PCA-RB \cite{deeba2020computer}           & 154          & 157         & 54          & 49.52          & 74.04          & 51.66          \\ 
OUS \cite{bernal2017comparative}                 & 131          & 57          & 77          & 69.7           & 63             & 66.1           \\ 
CUMED \cite{bernal2017comparative}                & 144          & 55          & 64          & 72.3           & 69.2           & 70.7           \\ 
Faster RCNN-based \cite{shin2018automatic}    & 167          & 26          & 41          & 81.5           & 80.3           & 80.9           \\ 
Mask RCNN-based \cite{qadir2019improving}     & N/A          & N/A         & N/A         & 80.0           & 72.59          & 76.12          \\ 
MDeNet-plus \cite{qadir2021toward}         & \textbf{180} & 28          & \textbf{28} & 86.12          & \textbf{86.54} & 86.33          \\
\textbf{COTR (Ours)} & 172          & \textbf{16} & 36          & \textbf{91.49} & 82.69          & \textbf{86.87} \\ \hline
\end{tabular}
\end{table}

\begin{table}[!t]
\centering
\caption{Performance comparison with state-of-the-art methods on CVC-ColonDB.}
\label{tab2}
\begin{tabular}{lllllll}
\hline
Method                        & TP  & FP  & FN & Pre   & Sen   & F1    \\ \hline
Discriminative feature-based \cite{bae2015polyp} & 212 & 88  & 88 & 70.67 & 70.67 & 70.67 \\ 
WE-SVM \cite{deeba2020computer}                       & 259 & 256 & 41 & 50.29 & 86.33 & 63.56 \\
MDeNetplus \cite{qadir2021toward}                   & 273 & 36  & 27 & 88.35 & 91    & 89.65 \\
\textbf{COTR (Ours)}                   & \textbf{275} & \textbf{19}  & \textbf{25} & \textbf{91.67} & \textbf{93.54} & \textbf{92.60} \\ \hline
\end{tabular}
\end{table}

\section{Conclusion}
\label{sec:conclusion}
In this paper, we proposed the COTR for end-to-end polyp detection. COTR embedded convolution layers into transformers for feature reconstruction and convergence acceleration. Experimental results on ETIS-LARIB and CVC-ColonDB dataset demonstrated that the proposed model achieved comparable performance against state-of-the-art methods. However, the COTR might produce low confidences when it encountered sessile polyps. We expect to address this issue in our future work.

%
%
%
\bibliographystyle{splncs04}
\bibliography{cotr_bib}

%




\end{document}